%% file: main.tex
\documentclass{article}
\usepackage{spconf,amsmath,graphicx,hyperref}

\usepackage{algorithm}
\usepackage{algorithmicx}
\usepackage{algpseudocode}

\usepackage{newfloat}
\usepackage{listings}

\usepackage{amssymb}

\usepackage{amsmath}
\usepackage{enumitem}
\usepackage{newfloat}
\usepackage{multirow}
\usepackage{makecell}
\usepackage{xcolor}

\usepackage{colortbl}
\usepackage{dsfont}
\usepackage{enumitem}
\usepackage{tcolorbox} 
\usepackage{mathrsfs}

\definecolor{myyellow}{rgb}{1,1, 0.6}
\definecolor{myorange}{rgb}{1, 0.8, 0.6}
\definecolor{myred}{rgb}{1, 0.6, 0.6}
\definecolor{second}{HTML}{FFDAB9}
\definecolor{best}{HTML}{FFC1C1}
\usepackage{multirow}
\usepackage{colortbl}
\usepackage{bbding}
\usepackage{stackrel}
\usepackage{booktabs}

\title{Enhancing Post-Training Quantization via Future Activation Awareness}
\name{Zheqi Lv$^{1*}$, Zhenxuan Fan$^{1*}$, Qi Tian$^{1,2}$, 
Wenqiao Zhang$^{1\dagger}$, Yueting Zhuang$^{1}$
\thanks{
$^{*}$These authors contributed equally to this work.\\
\indent\indent$^{\dagger}$Corresponding authors.
}}

\address{
\textsuperscript{$1$}Zhejiang University, Hangzhou, China\\
\textsuperscript{$2$}Tencent, Shenzhen, China
}

\newtheorem{theorem}{Theorem} 
\newcommand{\method}{FAQ}

\begin{document}
%
\maketitle
\input{tex/0abstract}
\input{tex/1introduction}

\input{tex/3method}

\input{tex/4experiment}
\input{tex/5conclusion}

\bibliographystyle{IEEEbib}
\bibliography{main}

\end{document}

%% file: tex/0abstract.tex
\begin{abstract}
\label{sec:abstract}
Post-training quantization (PTQ) is a widely used method to compress large language models (LLMs) without fine-tuning. It typically sets quantization hyperparameters (e.g., scaling factors) based on current-layer activations. 
Although this method is efficient, it suffers from quantization bias and error accumulation, resulting in suboptimal and unstable quantization, especially when the calibration data is biased.
To overcome these issues, we propose Future-Aware Quantization (FAQ), which leverages future-layer activations to guide quantization. This allows better identification and preservation of important weights, while reducing sensitivity to calibration noise. We further introduce a window-wise preview mechanism to softly aggregate multiple future-layer activations, mitigating over-reliance on any single layer.
To avoid expensive greedy search, we use a pre-searched configuration to minimize overhead. Experiments show that FAQ consistently outperforms prior methods with negligible extra cost, requiring no backward passes, data reconstruction, or tuning, making it well-suited for edge deployment.
\end{abstract}

%% file: tex/1introduction.tex
\section{Introduction}
\label{sec:introduction}

Recent years have witnessed rapid advancements in artificial intelligence, among which Large Language Models (LLMs) have achieved remarkable success across a variety of natural language processing tasks, including machine translation, question answering, and diverse downstream applications~\cite{bisk2020piqa,clark2018arc,zhang2024hyperllava,lv2025collaboration,yuan2025videorefer,zhang2022boostmis}.
However, the immense size of these models poses severe challenges for deployment on resource-constrained edge devices, which are increasingly expected to support private, low-latency, and offline inference.
To enable practical deployment, Post-Training Quantization (PTQ) methods have emerged as a promising direction for compressing pretrained models without requiring access to the original training data or performing costly retraining~\cite{banner2018post, nagel2020up, li2021brecq}.

Despite recent advances, most PTQ approaches rely on layer-wise quantization strategies that determine the quantization scaling factors based solely on the activation distribution of the current layer~\cite{wei2022outlier, yao2022zeroquant, dettmers2022llmint8}. This design introduces two critical issues:
(i) \textit{Quantization bias}. Channels that are crucial for downstream layers may be mistakenly compressed due to local decisions made at earlier layers. Specifically, outlier channels in the current layer can dominate the quantization range, suppressing other channels that may carry important information for subsequent computations. Conversely, some channels that are not essential to downstream performance may be preserved at higher precision at the expense of more important ones.
(ii) \textit{Error accumulation}. When relying solely on local activations, quantization behaves more like local optimization. Errors introduced in earlier layers propagate forward and accumulate across the network. These issues are further exacerbated when the calibration dataset has a distributional mismatch with real deployment data, significantly limiting the effectiveness of existing PTQ methods for deep and sensitive architectures like LLMs.

To address these challenges, we propose \textbf{\textit{Future-Aware Quantization (FAQ)}}, an activation-aware framework that leverages future-layer activations to guide the quantization process. Unlike traditional approaches that only consider local statistics, FAQ previews downstream activation distributions to assist in determining the quantization parameters for the current layer. This enables the retention of critical weights and allows the quantization process to be globally aligned with the model's forward sensitivity.
To further mitigate reliance on any single downstream layer and reduce sensitivity to potential future noise, we introduce a window-wise preview mechanism, which softly aggregates activations from multiple future layers.
To minimize computational overhead, we adopt a pre-searched configuration to eliminate the need for costly greedy hyperparameter search during calibration. We also provide a theoretical analysis that supports the effectiveness of FAQ.


\begin{figure*}[!th]
    \centering
    \includegraphics[width=\textwidth]{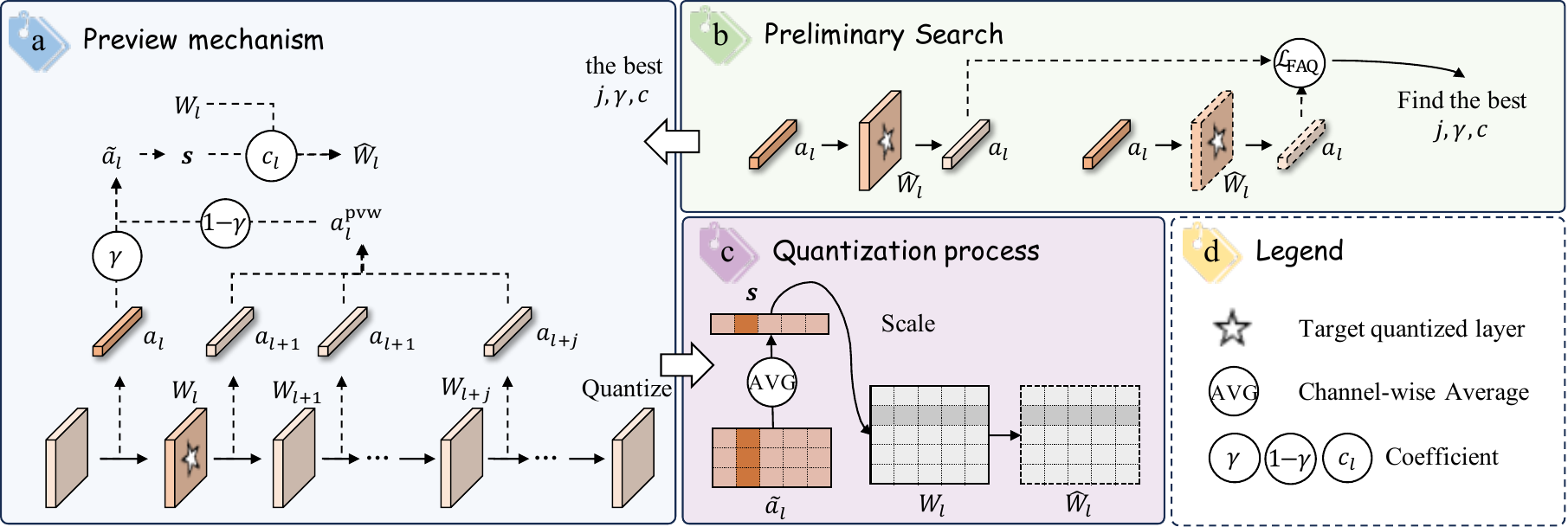}
    \vspace{-0.7cm}
    \caption{Overview of our proposed method \method{}.}
    \label{fig:method}
    \vspace{-0.3cm}
\end{figure*}

Our contributions are summarized as: (1) We reveal why current-layer-only PTQ suffers from bias and instability in deep LLMs. (2) We formulate FAQ, the first PTQ strategy that leverages future-layer activations, and devise a lightweight window-wise preview to balance accuracy and cost. Moreover, FAQ supports pre-searched configurations, significantly reducing computational overhead. (3) We provide a mathematical analysis to formally justify the effectiveness of our method. (4) Extensive experiments on several LLMs demonstrate that FAQ consistently surpasses strong PTQ baselines with almost zero additional computation or memory.

%% file: tex/3method.tex
\section{Methodology}
\label{sec:method}

\subsection{Preliminary and Notations}

Let $\mathcal{D}=\{x_i\}_{i=1}^{N}$ be an \emph{calibration set}. Each sample $x_i$ is a token sequence of length $T_i$; the maximum length is $T$. Batch size is $B$ in all forward passes. We set pretrained transformer LLM $\mathcal{M}$ has $L$ blocks. Layer $l$ owns weight $\mathbf{W}_i \in \mathbb{R}^{m \times n}$, where $m$ is the output dimension and $n$ is the input dimension. The activation input to the $\mathbf{W}_i$ is $\mathbf{a}_i$ and $\mathbf{a}_i\!\in\!\mathbb{R}^{B\times T\times m}$. For simplicity, we assume $B = 1$, so the activation reduces to: $\mathbf{a}_i \in \mathbb{R}^{T \times n}$. We compute the mean activation across the token dimension as: $\bar{\mathbf{a}}_i = \frac{1}{T} \sum_{t=1}^T \mathbf{a}_i^{(t)} \in \mathbb{R}^{1 \times n}$.


Post-Training Quantization (PTQ) compresses a pretrained full-precision model into a low-bit version using a calibration set $\mathcal{D}$ without backpropagation. It aims to quantize weights $\mathbf{W}_i$ for each layer $l$ while minimizing output degradation. We focus on weight-only PTQ where activations remain in full precision.

\noindent \textit{Quantization and Dequantization.}
Given bit-width $b$ (e.g., $b=4,8$), we define integer range $[-Q, Q{-}1]$ with $Q=2^{b-1}$. For a weight matrix $\mathbf{W} \in \mathbb{R}^{m \times n}$ and scale $s$, the symmetric quantizer is:
\begin{equation}
\mathcal{Q}(\mathbf{W} s) = \mathrm{clip}\left( \mathrm{round}(\mathbf{W} / \Delta), -Q, Q-1 \right) \cdot s.
\label{eq:q}
\end{equation}
This is equivalent to storing the quantized integer matrix $\mathbf{\hat{W}} = \mathrm{round}(\mathbf{W}/\Delta)$ and retrieving $W \approx \mathbf{\hat{W}} \cdot \Delta$ via dequantization during inference. Matrix multiplication is executed as INT × FP with rescaling. $\Delta$ is the quantization step size.

\noindent \textit{Base scale.}
The base scale $\mathbf{s}_i$ for layer $l$ is determined by a heuristic over activation $\mathbf{a}_i$. 
To scale the weight matrix according to the importance of each channel, AWQ considers the $k$-th row of $\mathbf{W}_i$ (denoted $\mathbf{w}_i^{(k)} \in \mathbb{R}^{1 \times n}$) and the $k$-th element of $\bar{\mathbf{a}}_i$, denoted $\bar{\mathbf{a}}_{i}^{(k)}$. Then set scale factor of the $i$-th layer $\mathbf{s}_i=\bar{a}_{i}$ and ${w}_i^{(k)} = \bar{a}_{i}^{(k)} \cdot {w}_i^{(k)}$

\noindent \textit{Loss and optimization.}
PTQ introduces a learnable multiplicative factor $c_i > 0$, and defines the effective quantization scale as $\mathbf{s}_i^* = c_i \cdot \mathbf{s}_i$. The quantized weight becomes:
\begin{equation}
\mathbf{\hat{W}}_i = \mathcal{Q}(\mathbf{W}_i, \mathbf{s}_i^*) = \mathcal{Q}(\mathbf{W}_i, s, c_i).
\end{equation}
We then minimize the following reconstruction loss:
\begin{equation}
\begin{aligned}
\mathcal{L}_{\text{PTQ}} &= \mathbb{E}_{x \sim \mathcal{D}} \left\| f(\mathbf{a}_i; \mathbf{\hat{W}}_i) - f(\mathbf{a}_i; \mathbf{W}_i) \right\|_2^2, \\
c_i^* &= \arg\min_{c_i} \mathcal{L}_{\text{PTQ}}.
\end{aligned}
\label{eq:ptq}
\end{equation}
This procedure is typically solved via grid search for $c_i$ using calibration data.

\noindent \textit{PTQ's Limitations.}
The base scale $\mathbf{s}_i$ depends only on current layer's statistics, causing:
(i) \emph{Error accumulation} as quantization errors propagate forward;
(ii) \emph{Quantization bias} where channels crucial for downstream layers are poorly preserved.

\subsection{Future-Aware Quantization}

To address the limitations of PTQ, we propose Future-Aware Quantization (FAQ). As shown in the Figure~\ref{fig:method}, the key idea is to adjust the scale computation by incorporating future-layer activations, thus aligning quantization with downstream sensitivity.


\noindent \textit{Layer-wise preview.}
Given a preview layer index difference $j \in \{1, \dots, N-i\}$, we then define the preview activation as $\mathbf{a}_i^{\mathrm{pvw}}=\mathbf{a}_{l+j}$

\noindent \textit{Window-wise preview.}
Given a preview window length $j \in \{1, \dots, N-i+1\}$, we define the preview activation as:
\begin{equation}
\mathbf{a}_i^{\mathrm{pvw}} = \frac{1}{j} \sum_{t=1}^{j} \mathbf{a}_{l+t}.
\end{equation}
After getting the preview activation, we then compute the fused activation:
\begin{equation}
\mathbf{\tilde{a}}_i = \gamma \cdot \mathbf{a}_i + (1 - \gamma) \cdot \mathbf{a}_i^{\mathrm{pvw}}, \quad \gamma \in (0, 1).
\label{eq:pvw}
\end{equation}
This fusion balances current-layer statistics with downstream context.

\noindent \textit{Future-aware base scale.}
We compute a new base scale using the fused activation $\mathbf{s}_i = \mathbf{\tilde{a}}_i$.
The effective quantization scale remains $\mathbf{s}_i^{*} = c_i \cdot \mathbf{s}_i$.

\noindent \textit{Loss and optimization.}
The quantized weight becomes:
\begin{equation}
\mathbf{\hat{W}}_i = \mathcal{Q}(\mathbf{W}_i, \mathbf{s}_i^*) = \mathcal{Q}(\mathbf{W}_i, \mathbf{s}, c_i,j,\gamma).
\end{equation}
Due to the change in $\mathbf{s}_i$, the loss form becomes:
\begin{equation}
\begin{aligned}
\mathcal{L}_{\text{FAQ}} = \mathbb{E}_{x \sim \mathcal{D}} 
\left\| f(\mathbf{a}_i; \mathcal{Q}(\mathbf{W}_i, \mathbf{s}_i^*)) - f(\mathbf{a}_i; \mathbf{W}_i) \right\|_2^2.
\end{aligned}
\label{eq:faq}
\end{equation}
And the optimization function is:
\begin{equation}
    (c_i^*, j^*, \gamma^*) = \arg\min_{c_i, j, \gamma} \mathcal{L}_{\text{FAQ}}.
\end{equation}
Thus, FAQ extends PTQ by generalizing the scale generation function.



\subsection{Theoretical Foundations of \method{}}

\begin{sloppypar}
\begin{theorem}
\label{theorem:1}
Given the following two assumptions: 
i) For the average magnitude of activation (per-channel) $\mathbf{a}_i=(a_i^{(1)},...,a_i^{(N)})$ of $N$ channels in the $i$-th layer, the value of the m-th channel is particularly large, that is, $\forall u,u\neq m,a_i^{(m)}\gg a_i^{(u)}$. Meanwhile, for the parameters $\mathbf{W}_l\left(l=i,i+1,...,I\right)$ of the $i$-th layer and its subsequent layers, the weight value $w_{l}^{jk}$ of the $(j,k)$-th position is significantly larger than that of other positions.
ii) Following AWQ~\cite{lin2023awq}, the larger the activation value of a channel, the larger the search range of $\mathbf{s}_{\mathbf{a}_i}^c=\left(\mathbf{a}_i\right)^c\left(c\in[0,1]\right)$ should be. 
Based on the above assumptions, it can be deduced that the quantization error of $\delta_{\mathrm{FAQ}}$ is smaller than the quantization error of $\delta_{\mathrm{AWQ}}$

\begin{align}
\delta_{\mathrm{FAQ}}&=\left\|\mathcal{Q}(f_{\mathrm{FAQ}}^1)f_{\mathrm{FAQ}}^2-\mathbf{a}_i\mathbf{W}_i\right\|_2 \nonumber\\
&<\left\|\mathcal{Q}(f_{\mathrm{AWQ}}^1)f_{\mathrm{AWQ}}^2-\mathbf{a}_i\mathbf{W}_i\right\|_2=\delta_{\mathrm{AWQ}},
\label{eq:theorem1}
\end{align}
where $f_{\mathrm{FAQ}}^1=\mathbf{W}_i\cdot\mathrm{diag}\left(\sum_{l=i}^I\gamma^l\left(\mathbf{a}_l\right)^c\right)$, $f_{\mathrm{FAQ}}^2=\mathrm{diag}\left(\sum_{l=i}^I\gamma^l\left(\mathbf{a}_l\right)^c\right)^{-1} \cdot \mathbf{a}_i$, $f_{\mathrm{AWQ}}^1=\mathbf{W}_i\cdot\mathrm{diag}\left(\left(\mathbf{a}_i\right)^c\right)$, $f_{\mathrm{AWQ}}^2=\mathrm{diag}\left(\left(\mathbf{a}_i\right)^c\right)^{-1}\cdot \mathbf{a}_i$ and $\gamma$ is the fusion factor that satisfied $\sum_{l=i}^I\gamma^l=1$.
\end{theorem}
\end{sloppypar}

%% file: tex/4experiment.tex
\section{Experiments}
\label{sec:experiments}

\subsection{Experimental Setup}

\input{tables/main_table}
We conduct evaluations of quantized pre-trained LLMs on both language generation and commonsense question answering tasks. In particular, we measure perplexity (\textit{PPL}) using the WikiText2~\cite{merity2016wiki} and C4~\cite{raffel2020c4} datasets, and assess zero-shot accuracy (\textit{Accuracy}) on the PIQA~\cite{bisk2020piqa}, ARC~\cite{clark2018arc}, BoolQ~\cite{clark2019boolq}, HellaSwag~\cite{zellers2019hellaswag}, and WinoGrande~\cite{sakaguchi2021winogrande} datasets.
To verify the applicability, we do experiments based on some widely used baselines including (a) LLMs. Qwen3 (4B,8B)~\cite{qwen3}, Qwen2.5 (0.5B, 7B)~\cite{qwen2.5}, LLaMA3.2 (3B)~\cite{touvron2024llama3}, LLaMA2 (7B)~\cite{touvron2023llama2}
. (b) Training-Free PTQ. round-to-nearest (RTN), AWQ~\cite{lin2023awq}.
We conducted experiments on an NVIDIA RTX 4090 GPU. Since \method{} involves hyperparameter tuning, we performed a preliminary search to fix the fusion factor $\gamma=0.85$ and window size $=3$ to reduce time cost. All results, except for the hyperparameter analysis experiments, follow this setting, though using a fully searched quantization strategy could yield better performance. The search strategy for the hyperparameter $\alpha$ is kept consistent with AWQ~\cite{lin2023awq}. We adopt asymmetric quantization.

\subsection{Experimental Results}
$\downarrow$ and $\uparrow$ indicate that lower and higher values are better, respectively. If a value is bolded in the results, it indicates the best performance.

\begin{sloppypar}
\noindent \textbf{Main Results.}
We evaluate \method{} at the 3-bit setting against full-precision (FP16) and popular PTQ baselines, including RTN and AWQ, across open-source LLMs: Qwen3-4B/8B, Qwen2.5-0.5B/7B, LLaMA2-7B and LLaMA3.2-3B
. Experiments cover language modeling and reasoning tasks, such as WikiText2, c4, arc\_challenge, hellaswag, winogrande, arc\_easy, boolq, and piqa, where lower perplexity and higher accuracy indicate better performance.
Table~\ref{tab:main_table} shows the results. Across benchmarks and models, \method{} consistently outperforms RTN and AWQ, proving its effectiveness. On Qwen3-8B, \method{} improves arc\_challenge accuracy from 0.4778 (AWQ) to 0.5043 and piqa from 0.7459 to 0.7535. On LLaMA-3.2B, \method{} achieves 0.3788 on arc\_challenge and 0.7563 on piqa, surpassing both RTN and AWQ. These results indicate that \method{}’s preview mechanism efficiently identifies future-sensitive channels, avoiding over-suppression common in local-only quantization strategies like RTN and AWQ.
\end{sloppypar}

\noindent \textbf{Impact of Model Size and Architecture.} \method{} generalizes well across both architecture and scale. For small models like Qwen2.5-0.5B, it reduces WikiText2 perplexity from 29.1318 (AWQ) to 25.9575 and improves boolq accuracy from 0.6171 to 0.6284. For 7B-scale models with varying backbones, such as Qwen2.5-7B, Qwen3-8B, and LLaMA2-7B, \method{} consistently improves over AWQ. On boolq, \method{} achieves 0.7840 (Qwen2.5-7B), 0.8529 (Qwen3-8B), and 0.7622 (LLaMA2-7B), all higher than AWQ. These consistent gains across families and sizes demonstrate \method{}’s scalability and architecture-agnostic nature.

\input{tables/main_table3}

\noindent \textbf{\method{} under 3bit vs. 4bit Settings.}
Table~\ref{tab:main_table3} focuses on aggressive quantization scenarios—evaluating 3bit and 4bit settings on boolq. We observe that \method{} shows larger improvements under 3bit, which presents more severe quantization noise and distortion:
On Qwen2.5-0.5B, the gain is +1.14 at 3bit, but disappears at 4bit.
This pattern reveals: at lower bit-widths, traditional methods suffer more from error accumulation and local quantization bias. \method{}’s forward-looking mechanism mitigates these effects, making it especially valuable for extreme low-bit scenarios. In summary, The relative advantage of \method{} is amplified at lower bit-widths, highlighting its importance in ultra-efficient quantization.

\input{tables/analysis_sample_num}

\subsubsection{The impact of calibrate dataset}

We analyze the robustness of \method{} against calibration data bias in Table~\ref{tab:calibrate_data}, in comparison to AWQ. We vary the value of $N$, where a smaller $N$ corresponds to greater bias in the calibration dataset, and a larger $N$ indicates less bias. As shown, \method{} consistently achieves better mean performance and lower variance across different sampling settings. This demonstrates a significant advantage of \method{} in mitigating calibration data sampling bias. The improvement is attributed to our preview mechanism, which enables the model to anticipate the influence of future layers and adapt accordingly, even when the activation distributions are diverse.

%% file: tables/main_table.tex
\begin{table*}[t]
\centering
\renewcommand{\arraystretch}{0.85}
\setlength{\tabcolsep}{7pt}
\resizebox{\textwidth}{!}{
\begin{tabular}{llcccccccc}
\toprule[2pt]
\textbf{LLM} & \textbf{Quant} & \textbf{wikitext2$\downarrow$} & \textbf{c4$\downarrow$} & \textbf{arc\_challenge$\uparrow$} & \textbf{hellaswag$\uparrow$} & \textbf{winogrande$\uparrow$} & \textbf{arc\_easy$\uparrow$} & \textbf{boolq$\uparrow$} & \textbf{piqa$\uparrow$} \\
\midrule
\multirow{4}{*}{Qwen3-4B} 
  & FP16 & 13.6372 & 16.7572 & 0.5068 & 0.5226 & 0.6614 & 0.8051 & 0.8511 & 0.7492 \\ \cmidrule{2-10}
  & RTN & 22.5701 & 27.4900 & 0.3490 & 0.4191 & 0.5604 & 0.6389 & 0.7502 & 0.6882
\\
  & AWQ  & 17.4229 & 20.2635 & 0.3925 & 0.4697 & 0.6164 & 0.6915 & \textbf{0.8000} & 0.7193 \\
  & \cellcolor{gray!10}\method{} (Ours) & \cellcolor{gray!10}\textbf{16.7608} & \cellcolor{gray!10}\textbf{20.0223} & \cellcolor{gray!10}\textbf{0.4087} & \cellcolor{gray!10}\textbf{0.4698} & \cellcolor{gray!10}\textbf{0.6259} & \cellcolor{gray!10}\textbf{0.7189} & \cellcolor{gray!10}0.7841 & \cellcolor{gray!10}\textbf{0.7236} \\
\midrule
\multirow{4}{*}{Qwen3-8B} 
  & FP16  & 9.7155	& 13.4575 & 0.5572 & 0.5715 & 0.6772 & 0.8359 & 0.8661 & 0.7693\\ \cmidrule{2-10}
  & RTN & 13.4763 & 17.8254 & 0.4556 & 0.4996 & 0.6085 & 0.7483 & 0.7948 & 0.7220\\
  & AWQ  & 11.6915 & 15.2169 & 0.4778 & \textbf{0.5296} & 0.6606 & 0.7870 & 0.8422 & 0.7459 \\
  & \cellcolor{gray!10}\method{} (Ours) & \cellcolor{gray!10}\textbf{11.5071} & \cellcolor{gray!10}\textbf{15.1996} & \cellcolor{gray!10}\textbf{0.5043} & \cellcolor{gray!10}0.5287 & \cellcolor{gray!10}\textbf{0.6772} & \cellcolor{gray!10}\textbf{0.8056} & \cellcolor{gray!10}\textbf{0.8529} & \cellcolor{gray!10}\textbf{0.7535} \\
\midrule
\multirow{4}{*}{LLaMA3.2-3B} 
  & FP16 & 7.8138 & 10.0394 & 0.4232 & 0.5529 & 0.7009 & 0.7449 & 0.7333 & 0.7677 \\ \cmidrule{2-10}
  & RTN & 13.2187 & 16.7545 & 0.3191 & 0.4735 & 0.6243 & 0.6254 & 0.6550 & 0.7263
 \\
  & AWQ  & 10.3003 & 13.5301 & 0.3558 & 0.5010 & 0.6685 & 0.6587 & \textbf{0.7263} & 0.7465 \\
  & \cellcolor{gray!10}\method{} (Ours) & \cellcolor{gray!10}\textbf{10.2469} & \cellcolor{gray!10}\textbf{13.5159} & \cellcolor{gray!10}\textbf{0.3788} & \cellcolor{gray!10}\textbf{0.5041} & \cellcolor{gray!10}\textbf{0.6732} & \cellcolor{gray!10}\textbf{0.6881} & \cellcolor{gray!10}0.7086 & \cellcolor{gray!10}\textbf{0.7563} \\

\midrule
\multirow{4}{*}{Qwen2.5-0.5B} 
  & FP16 & 13.0702 & 17.6278 & 0.2952 & 0.4062 & 0.5651 & 0.6452 & 0.6251 & 0.7013 \\ \cmidrule{2-10}
  & RTN & 50.2316 & 56.9807 & 0.2534 & 0.3278 & 0.4988 & 0.4899 & 0.5914 & 0.6251 \\
  & AWQ & 29.1318 & 32.5651 & 0.2662 & 0.3477 & \textbf{0.5422} & \textbf{0.5450} & 0.6171 & 0.6491 \\
  & \cellcolor{gray!10}\method{} (Ours) & \cellcolor{gray!10}\textbf{25.9575} & \cellcolor{gray!10}\textbf{30.8558} & \cellcolor{gray!10}\textbf{0.2389} & \cellcolor{gray!10}\textbf{0.3542} & \cellcolor{gray!10}0.5375 & \cellcolor{gray!10}0.5219 & \cellcolor{gray!10}\textbf{0.6284} & \cellcolor{gray!10}\textbf{0.6572} \\
\midrule
\multirow{4}{*}{Qwen2.5-7B} 
  & FP16 & 6.8486 & 10.6144 & 0.4770 & 0.6004 & 0.7293 & 0.8043 & 0.8471 & 0.7873 \\ \cmidrule{2-10}
  & RTN & 12.1092 & 16.3435 & 0.4189 & 0.5033 & 0.6519 & 0.7151 & 0.7771 & 0.7388 \\
  & AWQ & 8.1557 & 12.1342 & \textbf{0.4821} & 0.5587 & \textbf{0.6835} & \textbf{0.7950} & 0.8049 & 0.7704 \\
  & \cellcolor{gray!10}\method{} (Ours) & \cellcolor{gray!10}\textbf{8.0469} & \cellcolor{gray!10}\textbf{11.9269} & \cellcolor{gray!10}0.4522 & \cellcolor{gray!10}\textbf{0.5608} & \cellcolor{gray!10}0.6819 & \cellcolor{gray!10}0.7803 & \cellcolor{gray!10}\textbf{0.8330} & \cellcolor{gray!10}\textbf{0.7840} \\
\midrule
\multirow{4}{*}{LLaMA2-7B} 
  & FP16 & 5.4721 & 6.8420 & 0.3985 & 0.5669 & 0.6709 & 0.6928 & 0.7101 & 0.7835 \\ \cmidrule{2-10}
  & RTN & 6.6616 & 8.2004 & 0.3584 & 0.5466 & 0.6355 & 0.6742 & 0.6947 & 0.7612 \\
  & AWQ & 6.2438 & 7.6412 & \textbf{0.3959} & 0.5446 & 0.6496 & 0.6818 & 0.6685 & 0.7590 \\
  & \cellcolor{gray!10}\method{} (Ours) & \cellcolor{gray!10}\textbf{6.2191} & \cellcolor{gray!10}\textbf{7.6094} & \cellcolor{gray!10}0.3865 & \cellcolor{gray!10}\textbf{0.5447} & \cellcolor{gray!10}\textbf{0.6527} & \cellcolor{gray!10}\textbf{0.6528} & \cellcolor{gray!10}\textbf{0.6976} & \cellcolor{gray!10}\textbf{0.7622} \\
\bottomrule[2pt]
\end{tabular}
}
\vspace{-0.3cm}
\caption{
Perplexity ($\downarrow$) and Accuracy ($\uparrow$) results under weight-only quantization on various benchmarks.
}
\label{tab:main_table}
\vspace{-0.2cm}
\end{table*}

%% file: tables/main_table3.tex
\begin{table}[t]
\centering
\renewcommand{\arraystretch}{0.9}
\setlength{\tabcolsep}{15pt}
\resizebox{0.48\textwidth}{!}{
\begin{tabular}{llc|c}
\toprule[2pt]
\textbf{LLM} & \textbf{Quant} 
& \textbf{3bit$\uparrow$} 
& \textbf{4bit$\uparrow$} \\
\midrule
\multirow{4}{*}{Qwen2.5-0.5B}
  & FP16 & 0.6251 & 0.6251 \\ \cmidrule{2-4}
  & RTN  & 0.5914 & 0.6076 \\
  & AWQ  & 0.6171 & \textbf{0.6171} \\
  & \cellcolor{gray!10}\method{} (Ours) & \cellcolor{gray!10}\textbf{0.6284} & \cellcolor{gray!10}0.5422 \\
\midrule
\multirow{4}{*}{Qwen2.5-7B}
  & FP16 & 0.8471 & 0.8471 \\ \cmidrule{2-4}
  & RTN  & 0.7771 & 0.8385 \\
  & AWQ  & 0.8049 & 0.8040 \\
  & \cellcolor{gray!10}\method{} (Ours) & \cellcolor{gray!10}\textbf{0.8330} & \cellcolor{gray!10}\textbf{0.8180} \\
\bottomrule[2pt]
\end{tabular}
}
\vspace{-0.25cm}
\caption{
3bit and 4bit quantization results on \texttt{boolq}. 
}
\label{tab:main_table3}
\vspace{-0.4cm}
\end{table}

%% file: tables/analysis_sample_num.tex
\begin{table}[ht]
\centering
\setlength{\tabcolsep}{3pt}
\renewcommand{\arraystretch}{0.8}
\setlength{\tabcolsep}{8pt}
\resizebox{0.48\textwidth}{!}{
\begin{tabular}{c|c|c|c|c}
\toprule[2pt]
\textbf{Model} & \textbf{Method} & \textbf{N} & \textbf{wikitext2$\downarrow$} & \textbf{c4$\downarrow$} \\
\midrule
\multirow{12}{*}{Qwen2.5-7B}
  & \multirow{6}{*}{AWQ} & 16 & 8.0888 & 11.9657 \\
  &  & 32 & 8.2122 & 12.1902 \\
  &  & 64 & 8.0072 & 11.9076 \\
  &  & 128 & 8.1557 & 12.1342 \\
  \cmidrule{3-5}
  & & {Mean} & 8.1160 & 12.0494 \\
  & & {Std}  & 0.0883 & 0.1343 \\
  \cmidrule{2-5}
  & \multirow{6}{*}{\method{} (Ours)} & 16 & 8.0921 & 12.0251 \\
  &  & 32 & 8.0262 & 11.9383 \\
  &  & 64 & 8.0333 & 11.9384 \\
  &  & 128 & 8.0469 & 11.9269 \\
  \cmidrule{3-5}
  & & \cellcolor{gray!10}{Mean} & \cellcolor{gray!10}\textbf{8.0496} & \cellcolor{gray!10}\textbf{11.9572} \\
  & & \cellcolor{gray!10}{Std}  & \cellcolor{gray!10}\textbf{0.0296} & \cellcolor{gray!10}\textbf{0.0456} \\
\bottomrule[2pt]
\end{tabular}
}
\vspace{-0.25cm}
\caption{Comparison of AWQ and \method{} on Qwen2.5-7B under different $N$ (the number of the calibration data).}
\label{tab:calibrate_data}
\vspace{-0.3cm}
\end{table}

%% file: tex/5conclusion.tex
\section{Conclusion}
\label{sec:conclusion}

We present \textit{FAQ}, a lightweight PTQ method that uses future-layer activations to mitigate quantization bias and error accumulation. With a simple window-wise preview mechanism and pre-searched configuration, FAQ delivers consistent performance gains without backward passes or reconstruction. Its efficiency and generality suggest FAQ could support broader LLM deployment on edge devices, improving accessibility in compute- and memory-constrained environments.

\section{Acknowledgement}
\begin{sloppypar}   
This work has been supported in part by the NSFC
(No.62436007), the Key Research and Development Projects in Zhejiang Province (No.2025C01128, 2024C01106, 2025C01030, 2025C02156),
Ningbo Yongjiang Talent Introduction Programme (2023A400-G), Zhejiang University Education Foundation Qizhen Scholar Foundation.
\end{sloppypar}